\def\year{2021}\relax
\documentclass[letterpaper]{article} 
\usepackage{aaai21}  
\usepackage{times}  
\usepackage{helvet} 
\usepackage{courier}  
\usepackage[hyphens]{url}  
\usepackage{graphicx} 
\usepackage{natbib}  
\usepackage{caption} 
\usepackage{subfigure}
\usepackage{stfloats}
\usepackage{booktabs}
\usepackage{multirow}
\usepackage{multicol}
\usepackage{amsmath}
\usepackage{amsfonts}
\usepackage{algorithm}
\usepackage{algorithmic}
\usepackage{tabularx}
\usepackage{xcolor}
\usepackage{listings}
\usepackage[switch]{lineno}
\title{Slimmable Generative Adversarial Networks}
\author{
    Liang Hou,\textsuperscript{\rm 1}\textsuperscript{,\rm 2}\thanks{Work done as an intern at ByteDance AI Lab.}
    Zehuan Yuan,\textsuperscript{\rm 3}\thanks{Corresponding authors}
    Lei Huang,\textsuperscript{\rm 4}
    Huawei Shen,\textsuperscript{\rm 1}\textsuperscript{,\rm 2}\footnotemark[2]
    Xueqi Cheng,\textsuperscript{\rm 1}\textsuperscript{,\rm 2}
    Changhu Wang\textsuperscript{\rm 3}
    \\
}
\affiliations{

    \textsuperscript{\rm 1}CAS Key Laboratory of Network Data Science and Technology, \\Institute of Computing Technology, Chinese Academy of Sciences\\
    \textsuperscript{\rm 2}University of Chinese Academy of Sciences\\
    \textsuperscript{\rm 3}ByteDance AI Lab \\
    \textsuperscript{\rm 4}SKLSDE, Institute of Artificial Intelligence, Beihang University
    
    \{houliang17z, shenhuawei, cxq\}@ict.ac.cn, \{yuanzehuan, wangchanghu\}@bytedance.com, huanglei@nlsde.buaa.edu.cn
}

\begin{document}

\maketitle

\begin{abstract}
Generative adversarial networks (GANs) have achieved remarkable progress in recent years, but the continuously growing scale of models makes them challenging to deploy widely in practical applications.
In particular, for real-time generation tasks, different devices require generators of different sizes due to varying computing power.
In this paper, we introduce slimmable GANs (SlimGANs), which can flexibly switch the width of the generator to accommodate various quality-efficiency trade-offs at runtime.
Specifically, we leverage multiple discriminators that share partial parameters to train the slimmable generator.
To facilitate the \textit{consistency} between generators of different widths, we present a stepwise inplace distillation technique that encourages narrow generators to learn from wide ones.
As for class-conditional generation, we propose a sliceable conditional batch normalization that incorporates the label information into different widths.
Our methods are validated, both quantitatively and qualitatively, by extensive experiments and a detailed ablation study.
\end{abstract}

\section{Introduction}

One of the main reasons for the tremendous success of deep learning in recent years is the increasing scale of models.
In the branch of deep generative models, generative adversarial networks (GANs)~\cite{NIPS2014_5423} have received widespread attention and evolved from the original simple multi-layer perceptrons to the vast BigGAN framework~\cite{brock2018large} with residual blocks~\cite{He_2016_CVPR} and self-attention layers~\cite{pmlr-v97-zhang19d} to synthesize realistic images nowadays.
The arms race on increasing the model size is endless, while the computational power and budget of devices are limited, especially for mobile phones.
Several GAN applications such as photograph~\cite{Kupyn_2018_CVPR} and autonomous driving~\cite{zhang2018deeproad} require short response time and hopefully run on devices with limited computing power.
Recently, researchers began to develop lightweight GAN models.
However, different devices usually require customized models of different sizes to meet the given response time budget.
Moreover, even a single device needs models of different sizes due to several switchable performance modes, e.g., the high-performance mode and power-saving mode.
Consequently, numerous models need to be trained and deployed for a single task, which is also heavy work.

In this work, we are committed to developing a ``once-for-all'' generator, which we only train and deploy once but can flexibly switch the model size at runtime to address the practical challenges.
Inspired by slimmable neural networks (SNNs)~\cite{yu2018slimmable}, we focus on developing a generator with configurable widths, where the width refers to the number of channels in layers.
In addition to saving inference time, customization on width can reduce memory footprint during the layer-by-layer inference, while reducing depth cannot take this advantage.

Although several discriminative tasks such as image classification and object detection are well studied in SNNs, applying slimmable operators to GANs suffers from three following challenges:
First, how to accurately and appropriately estimate the divergence between generators at different widths and the real data through discriminators?
Second, how to ensure \textit{consistency} between generators of different widths? Here, the \textit{consistency} means that the generated images should be similar between these generators given the same latent code.
Third, how to incorporate the label information into generators at different widths in the class-conditional generation?

In this paper, we propose slimmable generative adversarial networks (SlimGAN) to combat the aforementioned problems.
First, we present discriminators with partially shared parameters to serve the generators at different widths.
Second, to improve the consistency between generators at different widths, we introduce a novel stepwise inplace distillation technique, which encourages narrow generators to learn from the wide generators.
Third, we propose a sliceable conditional batch normalization (scBN) to incorporate the label information into different widths on the basis of switchable batch normalization (sBN)~\cite{yu2018slimmable} for the class-conditional generation.
Extensive experiments across several real-world datasets and two neural network backbones demonstrate that SlimGAN can compete with or even outperform the individually trained GANs.
Remarkably, our proposed scBN achieves better performance with fewer parameters.
A systematic ablation study verifies the effectiveness of our design, including network framework and loss function.

\section{Related Work}

\subsection{Generative Adversarial Networks}

Generative adversarial networks (GANs)~\cite{NIPS2014_5423} were implemented by multi-layer perceptrons at the beginning.
To improve the capability of the generator and the discriminator, convolutional layers were introduced in DCGAN~\cite{radford2015unsupervised}.
Later, WGAN-gp~\cite{NIPS2017_7159} not only established flexible Lipschitz constraints but also brought the ResNet~\cite{He_2016_CVPR} backbone into the GAN literature. 
To further impose the Lipschitz constraint, SNGAN~\cite{miyato2018spectral} introduced spectral normalization to the discriminator, which is also applied to the generator in SAGAN~\cite{pmlr-v97-zhang19d}.
For class-conditional generation tasks, cGAN-pd~\cite{miyato2018cgans} injected the label information to the generator by employing conditional batch normalization (cBN)~\cite{NIPS2017_7237}, and the discriminator with projection technique.
Recently, BigGAN~\cite{brock2018large} was capable of generating diverse and realistic high-resolution images, mainly attributed to the massive model.

\subsection{Model Compression in GANs}

The arms race on developing increasingly bloated network architecture hinders the extensive deployment of GANs in practical applications.
To reduce the size of the generator, \citet{aguinaldo2019compressing} compressed GAN models using knowledge distillation techniques.
\citet{li2020gan} proposed a compression method for conditional GAN models.
Meanwhile, \citet{yu2020self} developed a self-supervised compression method that uses the trained discriminator to supervise the training of a compressed generator.
AutoGAN-Distiller~\cite{fu2020autogan} compressed GAN models using neural architecture search.
Recently, \citet{wang2020gan} developed a unified GAN compression framework, including model distillation, channel pruning, and quantization.

\subsection{Dynamic Neural Networks}

Unlike model compression, dynamic neural networks can adaptively choose the computational graph to reduce computation during training and inference.
For example, \citet{liu2018dynamic} presented an additional controller network to decide the computational graph depends on the input.
Similarly, \citet{hu2019learning} proposed to reduce the test time by introducing an early-exit gating function.
Different from adjusting the depth of neural networks, slimmable neural networks (SNNs)~\cite{yu2018slimmable} trained neural networks that can be executable at different widths, allowing immediate and adaptive accuracy-efficiency trade-offs at runtime.
Later, US-Net~\cite{Yu_2019_ICCV} extended SNN to universally slimmable scenarios and proposed improved training techniques.
AutoSlim~\cite{yu2019autoslim} utilized model pruning methods to obtain accuracy-latency optimal models but introduced additional storage consumption.
RS-Nets~\cite{wang2020resolution} proposed an approach to train neural networks which can switch image resolutions during inference.

Nevertheless, the aforementioned approaches are designed for discriminative tasks with a single neural network, while we focus on generative tasks based on GANs.
Since GAN consists of two networks, i.e., the generator and discriminator network, modifying the operational mechanism of the generator may destroy the stability of the entire system, which makes the training process of GAN with a slimmable generator challenging.

\section{Preliminaries}

\subsection{Generative Adversarial Networks}

Generative adversarial networks (GANs)~\cite{NIPS2014_5423} are typically composed of a generator and a discriminator.
Specifically, the generator $G:\mathcal{Z}\rightarrow\mathcal{X}$ learns to generate fake samples by mapping a random noise vector $z\in\mathcal{Z}$ in the latent space endowed with a predefined prior $\mathcal{P}_Z$ (e.g., multivariate normal distribution) to a sample $x\in\mathcal{X}$ in the high-dimensional complex data space.
The discriminator $D:\mathcal{X}\rightarrow [0,1]$ attempts to distinguish the synthetic examples generated by the generator from real data.
In contrast, the goal of the generator is to fool the discriminator by mimicking real data.
Formally, the objective function of GAN is formulated as follows:
\begin{equation}
\begin{aligned}
\min_{G}\max_{D} &\mathbb{E}_{x\sim \mathcal{P}_\text{data}}[\log(D(x))] +\\ 
&\mathbb{E}_{z\sim \mathcal{P}_Z}[\log(1-D(G(z)))],
\end{aligned}
\end{equation}
where $\mathcal{P}_\text{data}$ represents the underlying distribution of real data.
As proved in ~\cite{NIPS2014_5423}, this minimax game is considered as minimizing the Jansen Shannon (JS) divergence between the real data distribution and the generated one.
Ideally, the generator is supposed to converge until $\mathcal{P}_{G}=\mathcal{P}_\text{data}$.
The JS divergence estimated by the discriminator can be replaced with other $f$-divergences~\cite{NIPS2016_6066} or even true metrics such as Wasserstein distance~\cite{pmlr-v70-arjovsky17a} by modifying the objective function.

\subsection{Slimmable Neural Networks}
Slimmable neural networks (SNNs)~\cite{yu2018slimmable} can instantly adjust the network width according to the demands of various devices with different capacities.
Unlike other training lightweight model methods such as neural architecture search and model compression, SNN is more flexible because it only needs to be trained and deployed once to obtain multiple models at different widths from a pre-specified width list $\mathcal{W}$.
In order to avoid the discrepancy of mean and variance between networks at different widths, SNN proposed a switchable batch normalization (sBN), i.e., using independent BN learnable parameters for each width:
\begin{equation}
    x'_{w_i}=\gamma_{w_i}\frac{x_{w_i}-\mu(x_{w_i})}{\sigma(x_{w_i})}+\beta_{w_i},
\end{equation}
where $x_{w_i}$ represents the data batch at current width $w_i\in \mathcal{W}$.
Specifically, $\mu(\cdot)$ and $\sigma(\cdot)$ compute the mean and standard deviation of this batch, $\gamma_{w_i}$ and $\beta_{w_i}$ are learnable scale and shift, respectively, of the sBN at width $w_i$.

\begin{figure*}[ht]
\centering
\includegraphics[width=0.85\textwidth]{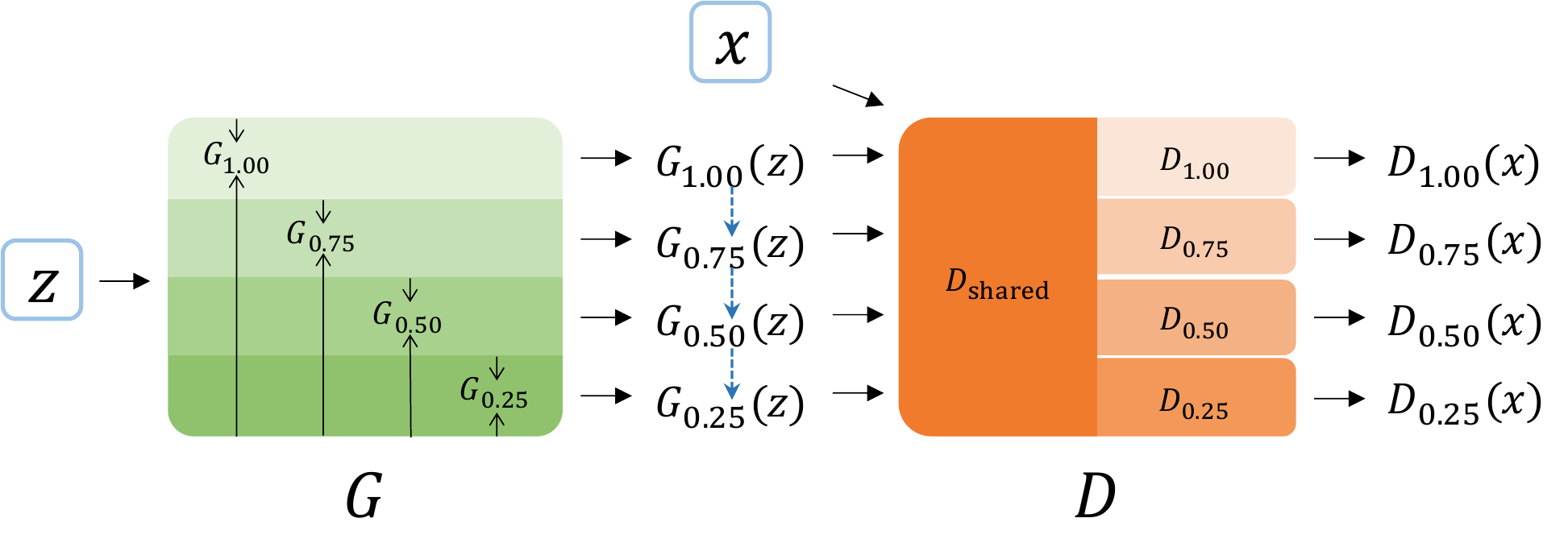}
\caption{Illustration of SlimGAN with width multiplier list $\mathcal{W}=[0.25, 0.5, 0.75, 1.0]\times$. Wide generators contain the channels of narrow ones. Multiple discriminators share the first several layers. Blue dashed lines indicate the stepwise inplace distillation.}
\label{model}
\end{figure*}

\section{Methods}

We aim to develop a size-flexible generator that can switch its size to accommodate various computing power.
Approximatively, the size-flexible generator implies multiple generators: $G_{\theta_1}, G_{\theta_2}, \cdots, G_{\theta_N}$ with $N$ incremental parameters $\theta_1 \subseteq \theta_2 \subseteq \cdots \subseteq \theta_N$, respectively.
In this work, we focus on slimming the width (number of channels) of the generator network instead of depth as reducing width can save memory footprint during the layer-by-layer inference.
The width-slimmable generator contains several generators: $G_{w_1}, G_{w_2}, \cdots, G_{w_N}$ at $N=|\mathcal{W}|$ incremental widths $w_1<w_2<...<w_N$ ($w_i\in \mathcal{W}$), respectively. 
Particularly, we train the generator via adversarial training and call our method slimmable GAN (SlimGAN)~\footnote{Code is available at \url{https://github.com/houliangict/SlimGAN}}.

\subsection{Slimmable GAN Framework}

We illustrate the overall framework of SlimGAN in Figure~\ref{model}.
Specifically, the SlimGAN consists of a slimmable generator with multi-width configurations and multiple discriminators that share the first several layers.
Each discriminator guides the generator at the corresponding width.
Here, using multiple shared discriminators, instead of a single discriminator or multiple independent discriminators, is critical for our SlimGAN model.
This is also the first major novelty of this model.
The idea is motivated by two insights.
On one hand, using a single discriminator for all the generators with different widths limits the flexibility and capability of discriminators to discriminate generated data from real data, and finally fails to obtain well-performed generators.
On the other hand, although assigning one discriminator for each generator offers high flexibility, it is incapable of leveraging the characteristic of data generated by slimmable generators.
Therefore, we borrow the idea of multi-task learning and design multiple parameter-shared discriminators.
This design not only offers high flexibility of discriminators but also leverages the similar characteristic of data generated by slimmable generators to improve the training of generators.
In addition, sharing parameters with other tasks offers a kind of consistency regularization on discriminators, which potentially improves the generalization of discriminators, and hence promotes the performance of generators~\cite{thanh-tung2018improving}.

As for training the generator-discriminator pair at width $w_i$, we utilize the Hinge version loss~\cite{lim2017geometric,tran2017deep}, which is prevalent and successful in GAN literature.
\begin{equation}
\begin{aligned}
    \max_{D} & \mathbb{E}_{x\sim \mathcal{P}_\text{data}}[\min(0,-1+D_{w_i}(x))] + \\ & \mathbb{E}_{z\sim \mathcal{P}_{Z}}[\min(0,-1-D_{w_i}(G_{w_i}(z)))] \\
    \max_{G} & \mathbb{E}_{z\sim \mathcal{P}_{Z}}[D_{w_i}(G_{w_i}(z))], i=1,2,\cdots,N
\end{aligned}
\end{equation}

\subsection{Stepwise Inplace Distillation}

Although a single slimmable generator implies multiple sub-generators, we expect these generators to maintain the \textit{consistency} between them, like an identical generator.
Imagine that, a trained slimmable generator is deployed as clients on various devices, and these devices may choose different width configurations according to their diverse energy budgets.
We expect these clients to generate consistent samples for the same command (e.g., the latent code $z$), which is broadcasted by the server.
We characterize this requirement as \textit{spatial translation consistency}.
In addition, since a single device has different performance modes, e.g., high-power mode or power-saving mode, even the same device may choose generators of different sizes.
We also expect this device to generate a consistent sample for the same latent code at any mode, which is considered as \textit{time translation consistency}.
However, the adversarial training objective function cannot explicitly guarantee the \textit{consistency} between generators of different widths because the discriminator only distinguishes real from fake but not distinguishes similar from dissimilar.

To achieve consistency, we propose a novel stepwise inplace distillation technique.
Different from general-purpose model distillation, we do not utilize knowledge distillation to obtain a smaller model through an already trained one. 
Instead, we train narrow networks by encouraging them to learn from wide networks during the training process, thereby improving consistency between them.
Specifically, the proposed distillation first distills the full generator to the second widest one and then distills the second one to the third one and so on.
We employ the pixel mean square error as the objective function in the distillation:
\begin{equation}
    \min_{G}\frac{\lambda}{N-1}\mathbb{E}_{z\sim\mathcal{P}_Z}\sum_{i=1}^{N-1}\|G_{w_i}(z)-\text{sg}(G_{w_{i+1}}(z))\|_2^2,
\end{equation}
where $\lambda$ is a hyper-parameter that balances the adversarial objectives and the distillation, and $\text{sg}(\cdot)$ means to stop the transfer of gradients in the computational graph. 
Stop updating the wide generator in distillation prevents it from learning from the narrow one.

Arguably, the distillation can effectively improve the performance of narrow networks.
Furthermore, the improvement of narrow networks could also lead to an enhancement of wide networks, because wide generators contain all the channels of narrow generators, which forms a virtuous cycle in SlimGAN.
As an alternative, leveraging the full network to teach all narrow generators, however, may be contrary to the assumption of width residuals~\cite{Yu_2019_ICCV}. 
In other words, forcing all narrow generators to learn from the widest one would make no difference between them, which may tend to strengthen the expression of parameters they shared but reduce the capability of their specific.

\subsection{Training Algorithm}

Algorithm~\ref{algorithm} shows the training procedure of SlimGAN in PyTorch-style pseudo-code. 
The main difference from training a normal GAN is that we enumerate all the widths in the pre-specified width list at each iteration and switch the computational graph according to the configured width.
In the adversarial training part, we sample random noises as the input of each generator.
This provides the diversity of fake samples, encouraging models to explore wider optimization space to achieve better results.
In the consistency training part, we sample the same latent code to optimize the discrepancy of the outputs between generators at different widths.

\begin{algorithm}[htbp]
\caption{Training SlimGAN} 
\label{algorithm}
\begin{algorithmic}[1]
\REQUIRE dataset $\mathcal{D}$, switchable width multiplier list $\mathcal{W}$
\ENSURE generator $G$
\FOR{$t=1,\dots,T$}
    \FOR{$k=1,\dots,K$}
        \STATE Get mini-batch data, $x=sample(\mathcal{D})$
        \FOR{$i=1,\dots,N$}
            \STATE Generate samples, $\hat{x}=G_{w_i}(z)$ with $z\sim\mathcal{P}_Z$
            \STATE Comp. D loss, $loss=lossD(D_{w_i}(x),D_{w_i}(\hat{x}))$
            \STATE Compute D gradients, $loss.backward()$
        \ENDFOR
        \STATE Update D weights, $optimizerD.step()$
    \ENDFOR
    \STATE Sample fixed noise $\bar{z}\sim \mathcal{P}_Z$ and initialize $\bar{x}=[\ ]$
    \FOR{$i=1,\dots,N$}
        \STATE Generate samples, $\hat{x}=G_{w_i}(z)$ with $z\sim\mathcal{P}_Z$
        \STATE Compute G loss, $loss=lossG(D_{w_i}(\hat{x}))$
        \STATE Compute G gradients, $loss.backward()$
        \STATE Generate fixed samples $\bar{x}.append(G_{w_i}(\bar{z}))$
    \ENDFOR
    \STATE Compute distillation loss, $loss=lossDistill(\bar{x})$
    \STATE Compute distillation gradients, $loss.backward()$
    \STATE Update G weights, $optimizerG.step()$
\ENDFOR
\RETURN $G$
\end{algorithmic}
\end{algorithm}

\subsection{Sliceable Conditional Batch Normalization}

In the case of class-conditional generation, state-of-the-art class-conditional GANs, e.g., BigGAN~\cite{brock2018large}, follow the way of incorporating label information proposed in cGAN-pd~\cite{miyato2018cgans}, i.e., conditional batch normalization (cBN) in the generator and projection in the discriminator.
In this work, we follow the label projection technique in the discriminator.
As for the generator, however, how to introduce the label information under the width-switchable mechanism is the key problem faced by SlimGAN in the class-conditional generation scenario.
In other words, how to unify sBN and cBN?
A naive way to achieve this goal is to expand each sBN to a cBN:
\begin{equation}
    x'_{w_i,c_j}=\gamma_{w_i,c_j}\frac{x_{w_i,c_j}-\mu(x_{w_i,c_j})}{\sigma(x_{w_i,c_j})}  + \beta_{w_i,c_j},
\end{equation}
where $c_j$ indicates the current label.
However, the disadvantages of this design are obvious from two perspectives.
First, the number of parameters increased dramatically because of $N\times C$ BN parameters ($C$ is the number of labels), which is contradictory to our motivation, i.e., saving parameters to reduce model size and computation.
Second, the information of the same label is separated for generators at different widths.

To remedy the above issues, we propose a sliceable conditional batch normalization (scBN) defined as follows:
\begin{equation}
    x'_{w_i,c_j}=\gamma_{w_i} \gamma_{c_j}^{:s_i} \frac{x_{w_i,c_j}-\mu(x_{w_i,c_j})}{\sigma(x_{w_i,c_j})}  + \beta_{w_i} + \beta_{c_j}^{:s_i},
\end{equation}
where $\gamma_{c_j}$ and $\beta_{c_j}$ are the learnable parameters of the cBN with label $c_j$.
To incorporate the label embedding into different widths, we slice cBN vectors to sub-vectors with the first $s_i=|\gamma_{w_i}|$ elements ($s_i$ is the number of channels in the layer at current width $w_i$).
Since cBN and sBN are independent, there are $N+C$ BN parameters in our proposed scBN, which not only accordingly reduces the parameters but also explicitly shares the information of the same label.

\section{Experiments}

In this section, we first evaluate our proposed SlimGAN across several datasets with two network backbones, compared with the individually trained models.
We then conduct class-conditional generation experiments to verify the effectiveness of scBN.
Besides, we report the qualitative and quantitative results that indicate the consistency between generators at different widths.
We further demonstrate the design of SlimGAN through an extensive ablation study.
We finally analyze the parameters complexities of generators.

\begin{table*}[ht]
\centering
\begin{tabular}{lllcccccccc}
\toprule
\multirow{2}*{Backbone} & \multirow{2}*{Dataset} & \multirow{2}*{Method} & \multicolumn{4}{c}{FID ($\downarrow$)} & \multicolumn{4}{c}{IS ($\uparrow$)} \\
& & & $0.25\times$ & $0.5\times$ & $0.75\times$ & $1.0\times$ & $0.25\times$ & $0.5\times$ & $0.75\times$ & $1.0\times$ \\

\midrule

\multirow{6}*{DCGAN (uncond)} & \multirow{2}*{CIFAR-10} & Individual & $46.9$ & $34.6$ & $30.4$ & $26.7$ & $6.08$ & $6.95$ & $7.39$ & $7.43$  \\
& & Slimmable & $\mathbf{37.3}$ & $\mathbf{28.5}$ & $\mathbf{25.8}$ & $\mathbf{25.2}$ & $\mathbf{6.90}$ & $\mathbf{7.31}$ & $\mathbf{7.43}$ & $\mathbf{7.44}$ \\
\cmidrule(lr){2-11}
& \multirow{2}*{STL-10} & Individual & $93.1$ & $69.1$ & $61.8$ & $57.4$ & $6.51$ & $7.82$ & $7.96$ & $8.38$ \\
& & Slimmable & $\mathbf{68.9}$ & $\mathbf{60.9}$ & $\mathbf{56.2}$ & $\mathbf{55.1}$ & $\mathbf{7.67}$ & $\mathbf{8.00}$ & $\mathbf{8.34}$ & $\mathbf{8.38}$ \\
\cmidrule(lr){2-11}
& \multirow{2}*{CelebA} & Individual & $24.4$ & $13.2$ & $10.4$ & $9.8$ & - & - & - & - \\
& & Slimmable & $\mathbf{23.3}$ & $13.3$ & $10.6$ & $\mathbf{9.4}$ & - & - & - & - \\

\midrule

\multirow{6}*{ResNet (uncond)} & \multirow{2}*{CIFAR-10} & Individual & $41.8$ & $24.1$ & $21.6$ & $20.3$ & $7.36$ & $7.68$ & $7.93$ & $7.91$ \\
& & Slimmable & $\mathbf{29.9}$ & $\mathbf{21.6}$ & $\mathbf{19.6}$ & $\mathbf{20.0}$ & $7.32$ & $\mathbf{8.02}$ & $\mathbf{8.15}$ & $\mathbf{8.09}$ \\
\cmidrule(lr){2-11}
& \multirow{2}*{STL-10} & Individual & $66.6$ & $58.5$ & $56.3$ & $52.9$ & $7.90$ & $8.52$ & $8.30$ & $8.60$ \\
& & Slimmable & $69.1$ & $59.0$ & $\mathbf{50.8}$ & $\mathbf{50.6}$ & $7.60$ & $8.23$ & $\mathbf{8.83}$ & $\mathbf{8.81}$ \\
\cmidrule(lr){2-11}
& \multirow{2}*{CelebA} & Individual & $18.0$ & $11.9$ & $9.9$ & $8.9$ & - & - & - & - \\
& & Slimmable & $\mathbf{13.9}$ & $\mathbf{10.6}$ & $\mathbf{9.8}$ & $\mathbf{8.5}$ & - & - & - & - \\

\midrule

\multirow{6}*{cGAN-pd (cond)} & \multirow{3}*{CIFAR-10} & Individual & $55.1$ & $33.5$ & $16.5$ & $15.5$ & $6.46$ & $7.90$ & $8.22$ & $8.52$ \\
& & Slimmable ($\times$) & $21.7$ & $17.2$ & $16.1$ & $16.2$ & $7.87$ & $8.31$ & $8.49$ & $8.34$ \\
& & Slimmable ($+$) & $\mathbf{19.5}$ & $\mathbf{14.5}$ & $\mathbf{13.6}$ & $\mathbf{14.2}$ & $\mathbf{7.88}$ & $\mathbf{8.38}$ & $\mathbf{8.67}$ & $\mathbf{8.59}$ \\
\cmidrule(lr){2-11}
& \multirow{3}*{CIFAR-100} & Individual & $45.8$ & $23.7$ & $22.5$ & $19.9$ & $7.26$ & $8.49$ & $8.50$ & $9.11$ \\
& & Slimmable ($\times$) & $26.8$ & $19.9$ & $18.9$ & $19.0$ & $8.13$ & $8.90$ & $9.14$ & $9.22$ \\
& & Slimmable ($+$) & $\mathbf{23.8}$ & $\mathbf{18.9}$ & $\mathbf{18.6}$ & $\mathbf{17.9}$ & $\mathbf{8.26}$ & $\mathbf{9.08}$ & $\mathbf{9.17}$ & $\mathbf{9.29}$ \\
\bottomrule
\end{tabular}
\caption{FID and IS on both unconditional (uncond) and class-conditional (cond) generation. 
We do not calculate IS on CelebA as it is a face dataset that lacking inter-class diversity, which IS measures. For class-conditional generation, ($+$) means our proposed sliceable conditional batch normalization while ($\times$) means the naive way that extends each sBN to cBN.
Bold numbers indicate our slimmable method outperforms the individually trained models.
}
\label{table_main}
\end{table*}

\subsection{Datasets}

We employ the following datasets for main experiments:
\textbf{CIFAR-10/100} consists of 50k training images and 10k validation images with resolution of $32\times32$. CIFAR-10 has 10 classes while CIFAR-100 has 100 classes.
\textbf{STL-10} is resized into the size of $48\times48$ as done in~\cite{miyato2018spectral}. There are 100k and 8k unlabeled images in the training set and validation set, respectively.
\textbf{CelebA} is a face dataset with 202,599 celebrity images with resolution of $178\times218$ originally. We follow the practice in~\cite{houdual} to center crop them to $178\times178$ and then resize them to $64\times64$. We divide the last 19,962 images into the validation set and the remaining 182,637 images as the training set.
We use the training set for training the models and the validation set for evaluation when calculating the statistics of the real data.

\subsection{Evaluation Metrics}

For evaluating the performance of all models on generation, we adopt two widely used evaluation metrics: Inception Score (IS)~\cite{NIPS2016_6125} and Fr\'echet Inception Distance (FID)~\cite{NIPS2017_7240}.
IS computes the KL divergence between the conditional class distribution and marginal class distribution.
FID is the Fr\'echet distance (the Wasserstein-2 distance between two Gaussian distributions) between two sets of features obtained through the Inception v3 network trained on ImageNet.
We randomly generate 50k images to calculate IS on all datasets, and 10k images to compute FID except STL-10, which we sample 8k images.

To measure the consistency between generators at different widths of SlimGAN, we present a metric, called Inception Consistency (IC), which measure the expected feature difference between two generators, $G_{w_i}$ and $G_{w_j}$ at width $w_i$ and $w_j$, respectively:
$$
    \text{IC}(G_{w_i}, G_{w_j})=\mathbb{E}_{z\sim \mathcal{P}_Z}[\|\Phi(G_{w_i}(z))-\Phi(G_{w_j}(z))\|_2^2],
$$
where $\Phi(\cdot)$ outputs the feature of the last hidden layer of Inception v3 network trained on ImageNet.

Given the width multiplier list $\mathcal{W}$, we average IC between all generator pairs as mean IC (mIC):
$$
    \text{mIC}(G, \mathcal{W})=\frac{1}{N\cdot (N-1)}\sum_{i=1}^{N}\sum_{j=1,i\neq j}^{N}\text{IC}(G_{w_{i}}, G_{w_{j}}).
$$
We randomly sample 10k images to estimate the mIC score.

\subsection{Experimental Settings}

We implement all models based on Mimicry~\cite{lee2020mimicry} using PyTorch framework.
The optimizer is Adam with betas $(\beta_1, \beta_2)=(0.5, 0.999)$ for DCGAN and $(\beta_1, \beta_2)=(0.0, 0.9)$ for ResNet based SNGAN.
The learning rate is $\alpha=2\times10^{-4}$, except CelebA on DCGAN, which is $\alpha=10^{-4}$.
The iterations of updating the generator are $T=100\text{k}$ for all methods.
The discriminator update steps per generator update step are $K=5$ for ResNet and $K=1$ for DCGAN.
As for the detailed network architecture, we exactly follow that in SNGAN~\cite{miyato2018spectral} and cGAN-pd~\cite{miyato2018cgans}.
The width multiplier list is set to $\mathcal{W}=[0.25, 0.5, 0.75, 1.0]\times$.

\subsection{Experimental Results}

\begin{figure*}[ht]
\centering

\subfigure[Slimmable GAN without the stepwise inplace distillation, showing clear inconsistency.]{
\label{figure2a}
\includegraphics[width=0.96\textwidth]{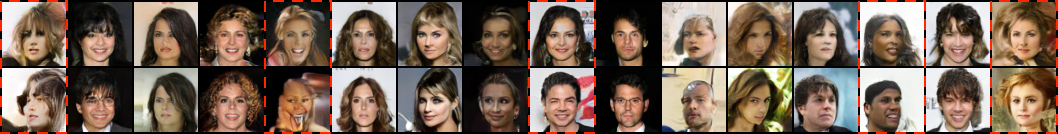}}
\subfigure[Slimmable GAN with the stepwise inplace distillation, showing improved consistency.]{
\label{figure2b}
\includegraphics[width=0.96\textwidth]{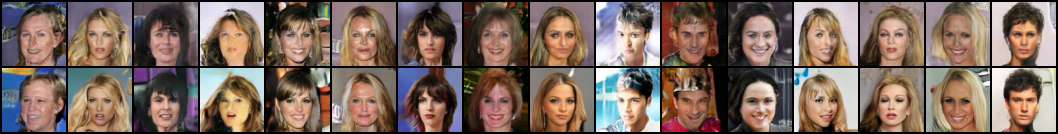}}
\caption{Qualitative consistency on CelebA.}
\label{figure2}
\end{figure*}

\subsubsection{Unconditional generation}

For unconditional generation, we experiment with three datasets, CIFAR-10, STL-10, and CelebA, on two backbones, DCGAN and ResNet.
The hyper-parameter is set as $\lambda=20$ for both backbones on CIFAR-10 and CelebA datasets, $\lambda=10$ and $\lambda=30$ for DCGAN and ResNet, respectively, on STL-10.
We report the FID and IS results in Table~\ref{table_main}.
Individual represents individually trained GANs of each width.
Our proposed SlimGAN surpasses in most cases or competes with the individually trained GANs in terms of both FID and IS scores, consistently demonstrating the effectiveness of SlimGAN across various datasets and network backbones.
Surprisingly, SlimGAN outperforms the individual model at the widest width.
We argue that the reasons are twofold. 
First, training narrow networks could provide extra informative signals for shared parameters with wide networks.
Second, the parameter-shared discriminators have a certain regularization, which may improve the generalization of each discriminator.
We believe this is a promising advanced training technique for GANs, and leave it for future work.
Additionally, some generators at width $0.75\times$ reach or surpass the widest generators, which are trained with only adversarial objectives, reflecting the benefit of the combination of distillation and adversarial training.

\subsubsection{Class-conditional generation}

For class-conditional generation experiments, we adopt cGAN-pd as the backbone on both CIFAR-10 and CIFAR-100, and report both FID and IS in the bottom of Table~\ref{table_main}. 
The hyper-parameter is set as $\lambda=10$ for CIFAR-10 and $\lambda=20$ for CIFAR-100.
The symbols in the parentheses after our slimmable methods represent different implementations of BN, i.e., $(\times)$ represents the naive BN, and $(+)$ represents our proposed scBN.
Overall, the slimmable generators with different BNs outperform the baseline heavily.
Particularly, our proposed scBN gains further improvement compared with the naive BN due to sharing the label information across different widths.

\begin{table}
\centering
\begin{tabular}{lcccc}

\toprule
\multirow{2}*{Methods} & \multicolumn{2}{c}{IS ($\uparrow$)} & \multicolumn{2}{c}{FID ($\downarrow$)} \\
& $0.5\times$ & $1.0\times$ & $0.5\times$ & $1.0\times$ \\
\midrule
Individual  & $18.8$ & $29.9$ & $48.1$ & $33.9$ \\
\midrule
Slimmable  & $\mathbf{32.7}$ & $\mathbf{36.1}$ & $\mathbf{32.8}$ & $\mathbf{30.8}$ \\
\bottomrule
\end{tabular}
\caption{BigGANs on ImageNet after 50k iterations.}
\label{table_biggan}
\vspace{-6px}
\end{table}

\subsubsection{BigGANs on ImageNet}

We train our slimmable method with BigGAN~\cite{brock2018large} on ImageNet ($128\times 128$) for 50k iterations.
The width multiplier list is set as $\mathcal{W}=[0.5, 1.0]\times$.
The IS and FID are reported in Table~\ref{table_biggan}.
In a word, our slimmable method surpasses the individually trained BigGANs, showing a strong capability on large-scale dataset of high-resolution images.

\subsubsection{Consistency}

\begin{table}[h]
\centering
\begin{tabular}{lccc}
\toprule
SlimDCGAN & CIFAR-10 & STL-10 & CelebA \\
\midrule
+ w/o distillation & $282.7$ & $277.4$ & $110.2$ \\
+ w/ distillation & $\mathbf{231.3}$ & $\mathbf{243.2}$ & $\mathbf{96.1}$ \\
\midrule
SlimResGAN & CIFAR-10 & STL-10 & CelebA \\
\midrule
+ w/o distillation & $285.7$ & $342.4$ & $116.9$ \\
+ w/ distillation & $\mathbf{241.4}$ & $\mathbf{248.7}$ & $\mathbf{97.9}$ \\
\bottomrule
\end{tabular}
\caption{mIC ($\downarrow$) on CIFAR-10, STL-10, and CelebA.}
\label{table_const}
\end{table}

We first report the quantitative consistency (mIC) in Table~\ref{table_const}, which verifies that distillation can improve the consistency.
We also show the qualitative consistency results on CelebA in Figure~\ref{figure2}.
For each method, the top row represents the narrowest generator and the bottom row indicates the widest generator.
The same column in each method shows the images generated through the same latent code.
Compared with the method without distillation, our distillation improves the consistency.
For example, the method without distillation synthesis faces with disparate hairs.

\subsection{Ablation Study}

\begin{table*}[ht]
\centering
\setlength{\tabcolsep}{4.25mm}{
\begin{tabular}{lcccccc}
\toprule
\multirow{2}*{DCGAN on CIFAR-10} & \multicolumn{5}{c}{FID ($\downarrow$)} & \multirow{2}*{mIC ($\downarrow$)} \\ 
& $0.25\times$ & $0.5\times$ & $0.75\times$ & $1.0\times$ & AVG  \\
\midrule
Individual & $46.9$ & $34.6$ & $30.4$ & $27.4$ & $34.8$ & - \\
Individual (full D) & $45.6$ & $33.2$ & $29.4$ & $27.4$ & $33.9$ & - \\
Slimmable G & $40.0$ & $35.2$ & $34.4$ & $33.4$ & $35.8$ & $264.3$ \\
+ shared D & $40.9$ & $30.2$ & $27.0$ & $25.2$ & $30.8$ &$282.7$ \\
+ shared D + distillation (SlimGAN) & $37.3$ & $\mathbf{28.5}$ & $\mathbf{25.8}$ & $\mathbf{25.2}$ & $\mathbf{29.2}$ & $231.3$ \\
\midrule
+ same D & $180.4$ & $136.9$ & $141.3$ & $158.6$ & $154.3$ & $376.8$ \\
+ slimmable D & $43.6$ & $35.8$ & $31.0$ & $33.0$ & $35.9$ & $269.5$ \\
+ distillation (w/o GAN loss for narrows) & $87.9$ & $56.2$ & $37.8$ & $28.9$ & $52.7$ & $\mathbf{204.8}$ \\
+ shared D + naive distillation & $\mathbf{36.6}$ & $29.8$ & $26.3$ & $25.5$ & $29.6$ & $232.5$ \\
\bottomrule
\end{tabular}}
\caption{Ablation Study on CIFAR-10. AVG means the averaged FID across all widths.}
\label{table_as}
\vskip -0.05in
\end{table*}

In this section, we conduct an extensive ablation study on CIFAR-10 to verify the effectiveness of the design in SlimGAN, including network framework and objective function.
The first two rows in Table~\ref{table_as} are both individually trained GANs.
Individual (full D) means the widths of all discriminators in these individual GANs are fixed as the widest width, which is consistent with SlimGAN.
Directly applying the slimmable operator to the generator with multiple independent discriminators (Slimmable G), unfortunately, obtains degradation, especially for wide generators.
Although this issue is alleviated by sharing partial parameters of these discriminators (shared D), it compromises consistency.
Fortunately, with stepwise inplace distillation, our final method (SlimGAN) not only achieves further improvements for narrow generators on generation but also obtains remarkable consistency.
When utilizing the same discriminator (same D) for all generators, the awful FID reveals that the one-to-one relationship in the generator-discriminator pair should be obeyed.
As an alternative parameter-sharing way, slimming the discriminator (slimmable D) does not gain satisfactory results.
This is because those narrow discriminators would lack the capability to estimate the divergences, as they are contained by wide discriminators.
Without adversarial training but only distillation for narrow generators, they tend to produce blurry images and get inferior FID.
Compared with the stepwise distillation, only the narrowest network is improved when using the naive distillation (all narrow generators learn from the widest one).

\subsection{Complexity Analysis}

\begin{table}[h]
\centering
\setlength{\tabcolsep}{1.85mm}{
\begin{tabular}{lccccc}
\toprule
CIFAR & $0.25\times$ & $0.5\times$ & $0.75\times$ & $1.0\times$ & Total \\
\midrule
I-uncond & $0.35$ & $1.15$ & $2.39$ & $4.08$ & $7.97$ \\
I-cond-10 & $0.36$ & $1.16$ & $2.41$ & $4.10$ & $8.04$ \\
I-cond-100 & $0.42$ & $1.29$ & $2.61$ & $4.37$ & $8.70$ \\
\midrule
S-uncond & - & - & - & - & $\mathbf{4.08}$ \\
S-cond-10 ($+$) & - & - & - & - & $\mathbf{4.11}$ \\
S-cond-100 ($+$) & - & - & - & - & $\mathbf{4.38}$ \\
\midrule
S-cond-10 ($\times$) & - & - & - & - & $4.15$ \\
S-cond-100 ($\times$) & - & - & - & - & $4.81$ \\
\bottomrule
\end{tabular}}
\caption{The number of parameters (M) in the generators.}
\label{table_params}
\end{table}

Saving parameters is the major advantage of the slimmable generator over the individually trained ones.
We investigate the number of parameters of unconditional (uncond) and class-conditional generators in Table~\ref{table_params}.
Specifically, cond10 and cond100 represent the class-conditional generators (cGAN-pd) that trained with 10 (CIFAR-10) and 100 (CIFAR-100) labels, respectively.
Individual (I-) methods require an independent generator on each width, while the slimmable (S-) approach only needs one.
Therefore, the slimmable generator reduces parameters greatly compared with the sum of all individuals.
As for class-conditional generative models, our proposed scBN ($+$) only adds negligible parameters on the widest individual generators compared to the naive BN approach.
This advantage would become more obvious with the increase of labels or switches.

\section{Conclusions}

In this paper, we introduce slimmable generative adversarial networks (SlimGAN), which can execute at different widths at runtime according to various energy budgets of different devices.
To this end, we utilize multiple discriminators that share partial parameters to train the slimmable generator.
In addition to the adversarial objectives, we introduce stepwise inplace distillation to explicitly guarantee the consistency between generators at different widths.
In the case of class-conditional generation, we propose a sliceable conditional batch normalization to incorporate the label information under the width-switchable mechanism.
Comprehensive experiments demonstrate that SlimGAN reaches or surpasses the individually trained GANs.
In the future, we will explore more practical generation tasks, e.g., text-to-image generation and image-to-image translation.

\section{ Acknowledgments}

This work is funded by the National Key R\&D Program of China (2020AAA0105200) and the National Natural Science Foundation of China under grant numbers 91746301 and U1911401. Huawei Shen is also funded by K.C. Wong Education Foundation and Beijing Academy of Artificial Intelligence (BAAI).

\bibliography{main}

\end{document}